\documentclass[runningheads]{llncs}
\usepackage{graphicx}

\usepackage{amssymb}

\usepackage{amsmath}
\usepackage[table,xcdraw]{xcolor}
\usepackage{multirow}
\usepackage{diagbox}
\usepackage{booktabs}
\usepackage[normalem]{ulem}
\useunder{\uline}{\ul}{}

\usepackage[misc]{ifsym} 
\begin{document}
\title{Equivalence of Correlation Filter and  Convolution Filter in Visual Tracking}
%
%
\author{Shuiwang Li\inst{2} \and
	Qijun Zhao\inst{1,2} \and
	Ziliang Feng\inst{1,2} \and
	Li Lu\inst{2} }
\authorrunning{Li et al.}
%
\institute{National Key Laboratory of Fundamental Science on Synthetic Vision, Sichuan University \and
	College of Computer Science, Sichuan University \\
	\email{lishuiwang0721@163.com, \{qjzhao,fengziliang,luli\}@scu.edu.cn}}
\maketitle              
\begin{abstract}
(Discriminative) Correlation Filter has been successfully applied to visual tracking and has advanced the field significantly in recent years. Correlation filter-based trackers consider visual tracking as a  problem of matching the feature template of the object and candidate regions in the detection sample, in which correlation filter provides the means to calculate the similarities. In contrast, convolution filter is usually used for blurring, sharpening, embossing, edge detection, etc in image processing. On the surface, correlation filter and convolution filter are usually used for different purposes. In this paper, however, we proves, for the first time, that correlation filter and convolution filter are equivalent in the sense that their minimum mean-square errors (MMSEs) in visual tracking are equal, under the condition that the optimal solutions exist and the ideal filter response is Gaussian and centrosymmetric. This result gives researchers the freedom to choose correlation or convolution in formulating their trackers. It also suggests that the explanation of the ideal response in terms of similarities is not essential.

\end{abstract}

\section{Introduction}
Visual tracking is a fundamental and challenging task in the field of computer vision, which has applications in numerous fields, e.g., video surveillance \cite{patel2012human}, disaster response \cite{2017Aerial}, intelligent traffic \cite{bota2011tracking} and wildlife protection \cite{2015Towards}, etc. However, visual tracking is still confronting with onerous challenges, e.g., object deformation, illumination variation, background clutter, motion blur, occlusion, visual angle and scale change, and real-time requirement, etc \cite{hu2011a}. 
Given its broad range of real-world applications, a number of large-scale benchmark datasets have been established, on which considerable methods have been proposed and demonstrated with significant progress in recent years. Two currently predominant approaches are discriminative correlation filter (DCF)-based methods and deep learning (DL)-based methods.
Deep learning has been intensively studied and demonstrated remarkable success in a wide range of computer vision areas, such as image classification, object detection, image caption and semantic segmentation, etc \cite{khan2020a,liu2017a,garcia-garcia2017a,zhao2019object,amirian2020automatic}.
Inspired by deep learning breakthroughs in these fields, DL-based methods have attracted considerable interest in the visual tracking community and witness rapid development and great advances in recent years.
And thanks to available large-scale datasets for training, DL-based trackers have achieved and are achieving state-of-the-art tracking performance and outperform DCF-based trackers significantly in terms of precision and accuracy.
Despite deep learning-based approaches have achieved great success and are promising in dealing with the challenges in visual tracking \cite{danelljan2020probabilistic,bhat2020know,li2019gradnet,wang2019fast}, its efficiency is unsatisfactory in the scenarios where computational resources is limited while real-time requirement is strict. One such real-world scenario is unmanned aerial vehicle (UAV)-based tracking where video sequences are captured by cameras mounted on board UAVs.
Equipped with visual tracking algorithms, UAV has
been wildly used in various applications, e.g., target following \cite{2017UAV} aircraft refueling \cite{yin2016robust}, disaster response \cite{2017Aerial}, autonomously landing \cite{2016Monocular}, and wildlife protection \cite{2015Towards}, etc. Compared with general tracking scenes, UAV tracking faces more onerous challenges \cite{2019Correlation}, in particular due to the limitations of onboard computing resources, battery capacity and maximum load of UAV, the deployment of DL-based tracking algorithms in UAV is still not feasible because deep neural networks usually need large storage space and computing resource, which are often difficult to meet in such small platforms as generic UAVs. DL-based tracking methods may be also not suitable for other samll platforms, for instance, mobile phone, handheld computer device and micro robot, etc, where resources are very limited. For this reason, DCF-base tracking algorithms because of their high CPU speed, which though are not as good as the DL-based tracking algorithms in terms of precision and accuracy, still attract a lot of attention of researchers and engineers, especially in the field of UAV tracking \cite{huang2019learning,li2020autotrack,li2020keyfilter,li2020training,li2019augmented,he2020towards}. 
That is why the study of DCF-based tracking algorithms still has important  value and application prospect. 

DCF-based trackers \cite{2015High,danelljan2017eco,2017Learning,li2018learning,li2020autotrack} are among the most efficient tracking algorithms for the time being \cite{2019Correlation}. 
These tracking algorithms consider object tracking as a problem of matching the feature template of the object and candidate regions in the detection sample, in which correlation filter provides the means to calculate the similarities. The candidate region with the highest similarity is usually taken as the new state of the object. According to the Parseval theorem and the correlation theorem, the correlation filters can be solved in the frequency domain efficiently and the correlation operation can be evaluated in an efficient way as well through the FFT (fast Fourier transform) algorithm. In contrast, convolution filter are usually used for blurring, sharpening, embossing, edge detection, etc in image processing \cite{pratt2013introduction}. On the surface, correlation filter and convolution filter are usually used for different purposes. In this paper, for the first time we proves that correlation filter and convolution filter are equivalent in the sense that their minimum mean-square error (MMSE) in visual tracking are equal, under the condition that the optimal solutions exist and the ideal filter response is Gaussian and centrosymmetric. Moreover, according to the Parseval theorem and the convolution theorem convolution filters can be solved in the frequency domain as well. This result gives researchers the freedom to choose correlation or convolution in formulating their trackers. It also suggests that the explanation of the ideal response in terms of similarities is not essential. In fact, current state-of-the-art DL-based methods are basically all Siamese trackers \cite{chen2020siamese,yu2020deformable,li2019siamrpn,li2018high}, which all employ a correlation operator to predict a target
confidence at each spatial position in a dense and efficient sliding-window manner in order to
localize the target, based upon which different heads may be used for either classification or regression or even other tasks. Therefore, our result also poses a question on whether the correlation operator could be replaced by the convolution operator and, furthermore, whether the similarity learning-based explanation of Siamese trackers is essential.

\section{Related Works}
The correlation filter first appeared in the field of signal processing, and later was applied to visual tracking, now known as the discriminative correlation filter (DCF). 
Below we provide a brief review on DCF-based trackers.
Bolme et al. \cite{bolme2010visual} first used correlation filter for visual tracking and proposed the minimum output sum of squared error (MOSSE) filter, which is considered the first DCF-based tracker. They considered the visual tracking as the problem of matching the initial object template and the candidate sub-images in the detection sample using correlation filter, and they proposed to efficiently solved the filter in the frequency domain. From the perspective of using cyclic sampling to obtain training samples for learning a linear regressor, the CSK tracker proposed by Henriques et al. \cite{henriques2012exploiting} actually obtained the same discriminant correlation filter, which were latter improved by the KCF \cite{2015High} tracker with kernel tricks and multi-channel features. By that time, DCF-based tracking algorithms not only showed good performance in various complex scenes, but also ran much faster than other types of tracking algorithms. Afterwards, DCF-based methods developed rapidly and increasingly achieved great improvements. For instance, to reduce boundary effects caused by the
periodic assumption of DCF Danelljan
et al. \cite{danelljan2015learning} proposed spatially regularized DCF (SRDCF) for tracking.
On the basis of SRDCF, Daneljan et al. further proposed the C-COT tracker \cite{danelljan2016beyond} and the ECO tracker \cite{danelljan2017eco} successively.
C-COT uses the VGG neural network to extract features and extends the feature maps of different resolutions to a continuous spatial domain through interpolation, and learns continuous
convolution operators for visual tracking.
The ECO tracker aims to simultaneously improve both speed and performance. It designs a factorized convolution operator which drastically reduces the number of parameters in the model, and a compact generative model of the training sample distribution that significantly reduces memory and time complexity. In addition, it uses a conservative model update strategy to improve robustness and to further reduce complexity.
Lukezic et al. \cite{lukezic2017discriminative} introduced the concept of channels and spatial reliabilities to DCF tracking in order to overcome the limitations related to the rectangular shape representation of the object. Mueller
et al. \cite{mueller2017context} proposed context-aware
DCF to incorporate global context to deal with fast motion, occlusion or background clutter on the grounds that
conventional DCF trackers are learned locally. By introducing temporal regularization to SRDCF, Li et al. \cite{li2018learning} proposed spatial-temporal regularized
correlation filters (STRCF) to achieve more robust appearance models. In order to online automatically and adaptively learn spatio-temporal regularization , Li et al. \cite{li2020autotrack} proposed an automatic spatio-Temporal regularization approach (AutoTrack). 
As for feature representation, the DCF-based approaches were initially restricted to a single feature channel and later extended to multi-channel feature maps \cite{Galoogahi2014Multi,danelljan2016beyond}, such as HOG \cite{Dalal2005Histograms} and color names \cite{Danelljan2014Adaptive} and deep CNN features \cite{Ma2015Hierarchical,danelljan2016beyond,sun2019roi,kart2019object}.
To efficiently model how both the foreground and background of the object varies over time, Galoogahi et al. \cite{2017Learning} proposed the backgroundaware correlation filter (BACF), which extracts patches densely from background using a cropping matrix and can be learned very efficiently. By enforcing
restriction to the rate of alteration in response maps generated in the detection phase, Huang et al. \cite{huang2019learning}, based upon the BACF tracker, proposed the aberrance repressed correlation filters (ARCF) to repress aberrances happening during the tracking process. Following BACF and ARCF, in order to improve the convergence properties and the wildly adopted discriminative scale estimation in DCF-based trackers, Li et al. \cite{li2021learning} proposed the residue-aware correlation filter (RACF) for UAV tracking in particular. BACF, ARCF and RACF all in fact can be thought of as asymmetric discriminative correlation filters proposed by Li et al. \cite{li2020asymmetric}, which have some theoretical advantages for visual tracking.
\section{Proof of the Equivalence of Correlation Filter and  Convolution Filter in Visual Tracking}
\subsection{Correlation Filter in Visual Tracking}
A DCF-based tracker aims to learn a multi-channel convolution filter $f$ from a set of training samples $\{x_k\}_{k=1}^{t}$, where each training sample $x_k$ consists of $d$ feature maps with $x_k^l$ denoting the $l_{th}$ one. The the coordinate of the maximum value of $y$ represents the center position of the object to be tracked. The filter $f$ consists of $d$ 2-D correlation filter $\{f^l\}_{l=1}^d$. Each $f^l$ and $x_k^l$ will be conducted the correlation operation, producing the correlation filter response. Let $x_k^l \in R^{m\times n}, f^l\in R^{m\times n}, N=mn$. Suppose $m$ and $n$ are even numbers to simplify our discussion. Then the correlation filter response of $f$
to $x_k$ is 
\begin{equation}
	\label{eq:correlation}
	R(x_k;f)=\sum_{l=1}^{d}x_k^l\circledcirc f^l,
\end{equation}
where $\circledcirc$ denotes the circular correlation operator.
The general objective function of correlation filter-based tracker is as follows,
\begin{equation}
	\label{eq:DCF}
	\mathop{\arg\min}_{f} \ \ \sum_{k=1}^{t}\alpha_k\| R(x_k;f)-y\|^2+\lambda\sum_{l=1}^{d}\|f^l\|^2,
\end{equation}
where $y$, defined by a 2-D centrosymmetric Gaussian function, is the ideal filter response, $t$ is the current frame number, $\alpha_k\geqslant 0$ decides the weight of each sample $x_k$, $\lambda$ is the penalty coefficient of the regularization. The matrix norm $\|\cdot\|$ is just the Frobenius norm, defined by
\begin{equation}\label{matrix_norm}
	\|f^l\|=\left ( \sum _{i=1}^m\sum _{j=1}^n|f(i,j)|^2\right )^{\frac{1}{2}}=\|\vec{{f}}^l\|, 
\end{equation}
where $\vec{f}^l$ denotes the vectorized $f^l$. It is worthy of note that the special case where $k=t$ and $\alpha_k=1$ is frequently used, when the objective function is more simple as follows \cite{fu2020correlation},  
\begin{equation}
	\label{eq:DCF_simplified}
	\mathop{\arg\min}_{f} \ \ \| R(x_k;f)-y\|^2+\lambda\sum_{l=1}^{d}\|f^l\|^2.
\end{equation}
Now, we discuss solving the problem (\ref{eq:DCF}). Substituting (\ref{eq:correlation}) into (\ref{eq:DCF}) we have
\begin{equation}
	\label{eq:DCF_rep}
	\mathop{\arg\min}_{f} \ \ \sum_{k=1}^{t}\alpha_k\| \sum_{l=1}^{d}x_k^l\circledcirc f^l-y\|^2+\lambda\sum_{l=1}^{d}\|f^l\|^2.
\end{equation}
According to the Parseval's theorem and the correlation theorem, (\ref{eq:DCF_rep}) is equivalent to
\begin{equation}
	\label{eq:DCF_solver_Eq1}
	\mathop{\arg\min}_{\hat{f}} \ \ \sum_{k=1}^{t}\alpha_k\| \sum_{l=1}^{d}\textup{conj}(\hat{x}_k^l)\odot \hat{f}^l-\hat{y}\|^2+\lambda\sum_{l=1}^{d}\|\hat{f}^l\|^2,
\end{equation}
where $\hat{z}$ denotes the Fourier transform of $z$, $\textup{conj}$ represents the operation of conjugate. It is obvious that (\ref{eq:DCF_solver_Eq1}) is equivalent to the following vectorized form,
\begin{equation}
	\label{eq:DCF_solver_Eq2}
	\mathop{\arg\min}_{\vec{\hat{f}}} \ \ \sum_{k=1}^{t}\alpha_k\| \sum_{l=1}^{d}\textup{conj}(\vec{\hat{x}}_k^l)\odot \vec{\hat{f}}^l-\vec{\hat{y}}\|^2+\lambda\sum_{l=1}^{d}\|\vec{\hat{f}}^l\|^2.
\end{equation}
Let $\hat{\mathbf{X}}_k=[\textup{diag}(\vec{\hat{x}}_k^1)^\textup{H},...,\textup{diag}(\vec{\hat{x}}_k^d)^\textup{H}]$, $\hat{\mathbf{f}}=[(\vec{\hat{f}}^1)^\textup{H},...,(\vec{\hat{f}}^d)^\textup{H}]^\textup{H}$, $\hat{\mathbf{y}}=\vec{\hat{y}}$, where $^\textup{H}$ represents the conjugate transpose. Then (\ref{eq:DCF_solver_Eq2}) is equivalent to 
\begin{equation}
	\label{eq:DCF_solver_Eq3}
	\mathop{\arg\min}_{\hat{\mathbf{f}}} \ \ \sum_{k=1}^{t}\alpha_k\| \hat{\mathbf{X}}_k\hat{\mathbf{f}}-\hat{\mathbf{y}}\|^2+\lambda\|\hat{\mathbf{f}}\|^2.
\end{equation}
This is a linear least square problem whose optimal solution $\hat{\mathbf{f}}_*$ (if exists) satisfies the following system of linear equations:
\begin{equation}
	\label{eq:DCF_solver_Eq4}
	[(\sum_{k=1}^{t}\alpha_k \hat{\mathbf{X}}_k^\textup{H}\hat{\mathbf{X}}_k)+\lambda \textup{I}_{dN} ]\hat{\mathbf{f}}_*=\sum_{k=1}^{t}\alpha_k \hat{\mathbf{X}}_k^\textup{H}\hat{\mathbf{y}},
\end{equation}
where $\textup{I}_{dN}$ is a identity matrix of size  $dN\times dN$. If $[(\sum_{k=1}^{t}\alpha_k \hat{\mathbf{X}}_k^\textup{H}\hat{\mathbf{X}}_k)+\lambda \textup{I}_{dN} ]$ is invertible, then
\begin{equation}
	\label{eq:DCF_solver_Eq5}
	\hat{\mathbf{f}}_*=[(\sum_{k=1}^{t}\alpha_k \hat{\mathbf{X}}_k^\textup{H}\hat{\mathbf{X}}_k)+\lambda \textup{I}_{dN} ]^{-1}(\sum_{k=1}^{t}\alpha_k \hat{\mathbf{X}}_k^\textup{H}\hat{\mathbf{y}}).
\end{equation}
When $d=1$, $[(\sum_{k=1}^{t}\alpha_k \hat{\mathbf{X}}_k^\textup{H}\hat{\mathbf{X}}_k)+\lambda \textup{I}_{dN} ]$ reduces to a diagonal matrix. If it is invertible, $\hat{\mathbf{f}}$ can be easily obtained. But when $d\neq 1$, the inversion of $[(\sum_{k=1}^{t}\alpha_k \hat{\mathbf{X}}_k^\textup{H}\hat{\mathbf{X}}_k)+\lambda \textup{I}_{dN} ]$ becomes much complicated. In fact, Li et al. \cite{li2020asymmetric} proved, under general conditions, that $[(\sum_{k=1}^{t}\alpha_k \hat{\mathbf{X}}_k^\textup{H}\hat{\mathbf{X}}_k)+\lambda \textup{I}_{dN} ]$ is a block matrix with each block is a diagonal matrix if $d\neq 1$. 

\subsection{Convolution Filter in Visual Tracking}

In this paper, the filter constructed by the convolution operator is called the convolution filter to distinguish it from the correlation filter, although no distinction is made between them in some literature \cite{danelljan2015learning,danelljan2016beyond,danelljan2017eco}.
Replacing the circular correlation operator in the objective function (\ref{eq:DCF_rep}) by the circular convolution operator results in the objective function of convolution filter-based tracker as follows, 
\begin{equation}
	\label{Equv_conv_corre_Eq1}
	\mathop{\arg\min}_{f} \ \ \sum_{k=1}^{t}\alpha_k\| \sum_{l=1}^{d}x_k^l\circledast f^l-y\|^2+\lambda\sum_{l=1}^{d}\|f^l\|^2.
\end{equation}
According to the Parseval theorem and the convolution theorem, (\ref{Equv_conv_corre_Eq1}) is equivalent to
\begin{equation}
	\label{Equv_conv_corre_Eq2}
	\mathop{\arg\min}_{\hat{f}} \ \ \sum_{k=1}^{t}\alpha_k\| \sum_{l=1}^{d}\hat{x}_k^l\odot \hat{f}^l-\hat{y}\|^2+\lambda\sum_{l=1}^{d}\|\hat{f}^l\|^2.
\end{equation}
And (\ref{Equv_conv_corre_Eq2}) is obviously equivalent to the following vectorized formulation:
\begin{equation}
	\label{Equv_conv_corre_Eq3}
	\mathop{\arg\min}_{\vec{\hat{f}}} \ \ \sum_{k=1}^{t}\alpha_k\| \sum_{l=1}^{d}\vec{\hat{x}}_k^l\odot \vec{\hat{f}}^l-\vec{\hat{y}}\|^2+\lambda\sum_{l=1}^{d}\|\vec{\hat{f}}^l\|^2.
\end{equation}
Let $\hat{\mathbf{X'}}_k=[\textup{diag}(\vec{\hat{x}}_k^1),...,\textup{diag}(\vec{\hat{x}}_k^d)]$, $\hat{\mathbf{f'}}=[(\vec{\hat{f}}^1)^\textup{H},...,(\vec{\hat{f}}^d)^\textup{H}]^\textup{H}$, $\hat{\mathbf{y}}=\vec{\hat{y}}$, then (\ref{Equv_conv_corre_Eq2}) is equivalent to 
\begin{equation}
	\label{Equv_conv_corre_Eq4}
	\mathop{\arg\min}_{\hat{\mathbf{f'}}} \ \ \sum_{k=1}^{t}\alpha_k\| \hat{\mathbf{X'}}_k\hat{\mathbf{f'}}-\hat{\mathbf{y}}\|^2+\lambda\|\hat{\mathbf{f'}}\|^2.
\end{equation}
The optimal solution $\hat{\mathbf{f'}}_*$ of (\ref{Equv_conv_corre_Eq4}) (if exists) satisfies the following system of linear equations:
\begin{equation}
	\label{Equv_conv_corre_Eq5}
	[(\sum_{k=1}^{t}\alpha_k \hat{\mathbf{X'}}_k^\textup{H}\hat{\mathbf{X'}}_k)+\lambda \textup{I}_{dN} ]\hat{\mathbf{f'}}_*=\sum_{k=1}^{t}\alpha_k \hat{\mathbf{X'}}_k^\textup{H}\hat{\mathbf{y}}.
\end{equation}
If $[(\sum_{k=1}^{t}\alpha_k \hat{\mathbf{X'}}_k^\textup{H}\hat{\mathbf{X'}}_k)+\lambda \textup{I}_{dN} ]$ is invertible, then
\begin{equation}
	\label{Equv_conv_corre_Eq6}
	\hat{\mathbf{f'}}_*=[(\sum_{k=1}^{t}\alpha_k \hat{\mathbf{X'}}_k^\textup{H}\hat{\mathbf{X'}}_k)+\lambda \textup{I}_{dN} ]^{-1}(\sum_{k=1}^{t}\alpha_k \hat{\mathbf{X'}}_k^\textup{H}\hat{\mathbf{y}}).
\end{equation}
\subsection{Proof of the Equivalence}

Let
\begin{equation}\label{Equv_conv_corre_Eq7}
	\begin{split}
		&\hat{\mathbf{s}}(\{\hat{\mathbf{G}}_k(\vec{\hat{x}}_k^1,...,\vec{\hat{x}}_k^d)\}_{k=1}^t;\{\alpha_k\}_{k=1}^t,\hat{\mathbf{y}})\\
		&=[(\sum_{k=1}^{t}\alpha_k \hat{\mathbf{G}}_k^\textup{H}(\vec{\hat{x}}_k^1,...,\vec{\hat{x}}_k^d)\hat{\mathbf{G}}_k(\vec{\hat{x}}_k^1,...,\vec{\hat{x}}_k^d))+\lambda \textup{I}_{dN} ]^{-1}(\sum_{k=1}^{t}\alpha_k \hat{\mathbf{G}}_k^\textup{H}(\vec{\hat{x}}_k^1,...,\vec{\hat{x}}_k^d)\hat{\mathbf{y}}),\\
		&\hat{\mathbf{G}}_k(\vec{\hat{x}}_k^1,...,\vec{\hat{x}}_k^d)=[\textup{diag}(\vec{\hat{x}}_k^1),...,\textup{diag}(\vec{\hat{x}}_k^d)].
	\end{split}
\end{equation}
If the optimal solution of the correlation filter-based tracker defined by (\ref{eq:DCF_solver_Eq2}) and that of the convolution filter-based tracker defined by (\ref{Equv_conv_corre_Eq2}) are considered as functions of $\{\hat{x}_k\}_{k=1}^t$, then the optimal solutions given in (\ref{eq:DCF_solver_Eq5}) and (\ref{Equv_conv_corre_Eq6}) respectively satisfy the following equations:
\begin{equation}
	\begin{split}
		&\hat{\mathbf{f}}_*(\vec{\hat{x}}_k^1,...,\vec{\hat{x}}_k^d)=\hat{\mathbf{s}}(\{\hat{\mathbf{G}}_k(\textup{conj}(\vec{\hat{x}}_k^1),...,\textup{conj}(\vec{\hat{x}}_k^d))\}_{k=1}^t;\{\alpha_k\}_{k=1}^t,\hat{\mathbf{y}})\\
		&\hat{\mathbf{f'}}_*(\vec{\hat{x}}_k^1,...,\vec{\hat{x}}_k^d)=\hat{\mathbf{s}}(\{\hat{\mathbf{G}}_k(\vec{\hat{x}}_k^1,...,\vec{\hat{x}}_k^d)\}_{k=1}^t;\{\alpha_k\}_{k=1}^t,\hat{\mathbf{y}}).
	\end{split}
	\label{Equv_conv_corre_Eq8}
\end{equation}
Therefore, it follows that
\begin{equation}\label{Equv_conv_corre_Eq9}
	\begin{split}
		&\hat{\mathbf{f}}_*(\{\vec{\hat{x}}_k^1,...,\vec{\hat{x}}_k^d\}_{k=1}^t)=\hat{\mathbf{f'}}_*(\{\textup{conj}(\vec{\hat{x}}_k^1),...,\textup{conj}(\vec{\hat{x}}_k^d)\}_{k=1}^t)\\
		&\hat{\mathbf{f'}}_*(\{\vec{\hat{x}}_k^1,...,\vec{\hat{x}}_k^d\}_{k=1}^t)=\hat{\mathbf{f}}_*(\{\textup{conj}(\vec{\hat{x}}_k^1),...,\textup{conj}(\vec{\hat{x}}_k^d)\}_{k=1}^t).
	\end{split}
\end{equation}
Since $y$ is a 2-D centrosymmetric Gaussian function, the Fourier transform of $y$ is also a Gaussian function. Therefore, $\hat{\mathbf{y}}$ is real valued. So we have $\textup{conj}(\hat{\mathbf{y}})=\hat{\mathbf{y}}$. Thus, 
\begin{equation}\label{Equv_conv_corre_Eq10}
	\begin{split}
		&\textup{conj}(\hat{\mathbf{f'}}_*(\{\vec{\hat{x}}_k^1,...,\vec{\hat{x}}_k^d\}_{k=1}^t))=\textup{conj}(\hat{\mathbf{s}}(\{\hat{\mathbf{G}}_k(\vec{\hat{x}}_k^1,...,\vec{\hat{x}}_k^d)\}_{k=1}^t;\{\alpha_k\}_{k=1}^t,\hat{\mathbf{y}}))\\
		&=\textup{conj}([(\sum_{k=1}^{t}\alpha_k \hat{\mathbf{G}}_k^\textup{H}(\vec{\hat{x}}_k^1,...,\vec{\hat{x}}_k^d)\hat{\mathbf{G}}_k(\vec{\hat{x}}_k^1,...,\vec{\hat{x}}_k^d))+\lambda \textup{I}_{dN} ]^{-1}(\sum_{k=1}^{t}\alpha_k \hat{\mathbf{G}}_k^\textup{H}(\vec{\hat{x}}_k^1,...,\vec{\hat{x}}_k^d)\hat{\mathbf{y}}))\\
		&=[(\sum_{k=1}^{t}\alpha_k \hat{\mathbf{G}}_k^\textup{H}(\textup{conj}(\vec{\hat{x}}_k^1),...,\textup{conj}(\vec{\hat{x}}_k^d))\hat{\mathbf{G}}_k(\textup{conj}(\vec{\hat{x}}_k^1),...,\textup{conj}(\vec{\hat{x}}_k^d)))+\lambda \textup{I}_{dN} ]^{-1}\\
		&\quad \quad (\sum_{k=1}^{t}\alpha_k \hat{\mathbf{G}}_k^\textup{H}(\textup{conj}(\vec{\hat{x}}_k^1),...,\textup{conj}(\vec{\hat{x}}_k^d))\hat{\mathbf{y}})\\
		&=\hat{\mathbf{f'}}_*(\{\textup{conj}(\vec{\hat{x}}_k^1),...,\textup{conj}(\vec{\hat{x}}_k^d)\}_{k=1}^t)\\
		&=\hat{\mathbf{f}}_*(\{\vec{\hat{x}}_k^1,...,\vec{\hat{x}}_k^d\}_{k=1}^t).
	\end{split}
\end{equation}
Note that $\textup{conj}(\hat{\mathbf{G}}_k(\vec{\hat{x}}_k^1,...,\vec{\hat{x}}_k^d))=\hat{\mathbf{G}}_k(\textup{conj}(\vec{\hat{x}}_k^1),...,\textup{conj}(\vec{\hat{x}}_k^d))$ was used in the derivation of (\ref{Equv_conv_corre_Eq10}).
Denote the time domain expressions of the optimal solutions $\hat{\mathbf{f}}_*$ and $\hat{\mathbf{f'}}_*$ by $f_*$ and $f_*'$ respectively. For a given detection sample $x_{k'}$ the filer responses corresponding to $f_*$ and $f_*'$, respectively, are $R(x_{k'};f_*)=\sum_{l=1}^{d}x_{k'}^l\circledcirc f_*^l$ and $R'(x_{k'};f_*')=\sum_{l=1}^{d}x_{k'}^l\circledast {f'}_*^{l}$.
Since $\hat{\mathbf{X}}_{k'}(\vec{\hat{x}}_{k'}^1,...,\vec{\hat{x}}_{k'}^d)=[\textup{diag}(\vec{\hat{x}}_{k'}^1)^\textup{H},...,\textup{diag}(\vec{\hat{x}}_{k'}^d)^\textup{H}]$ and $\hat{\mathbf{X'}}_{k'}(\vec{\hat{x}}_{k'}^1,...,\vec{\hat{x}}_{k'}^d)=[\textup{diag}(\vec{\hat{x}}_{k'}^1),...,\textup{diag}(\vec{\hat{x}}_{k'}^d)]$, it follows that
\begin{equation}\label{Equv_conv_corre_Eq10_1}
	\hat{\mathbf{X}}_{k'}(\vec{\hat{x}}_{k'}^1,...,\vec{\hat{x}}_{k'}^d)=\textup{conj}(\hat{\mathbf{X'}}_{k'}(\vec{\hat{x}}_{k'}^1,...,\vec{\hat{x}}_{k'}^d)).
\end{equation}
Therefore,
\begin{equation}\label{Equv_conv_corre_Eq11}
	\begin{split}
		&R(x_{k'};f_*)[i,j]=(\sum_{l=1}^{d}x_{k'}^l\circledcirc f_*^l)[i,j]\\
		&=\textup{F}^{-1}\{\textup{vec}^{-1}[\hat{\mathbf{X}}_{k'}(\vec{\hat{x}}_{k'}^1,...,\vec{\hat{x}}_{k'}^d)\hat{\mathbf{f}}_*(\{\vec{\hat{x}}_k^1,...,\vec{\hat{x}}_k^d\}_{k=1}^t)]\}[i,j]\\
		&=\textup{F}^{-1}\{\textup{vec}^{-1}[\textup{conj}(\hat{\mathbf{X'}}_{k'}(\vec{\hat{x}}_{k'}^1,...,\vec{\hat{x}}_{k'}^d))\textup{conj}(\hat{\mathbf{f'}}_*(\{\vec{\hat{x}}_k^1,...,\vec{\hat{x}}_k^d\}_{k=1}^t))]\}[i,j]\\
		&=\textup{F}^{-1}\{\textup{vec}^{-1}[\textup{conj}\left(\hat{\mathbf{X'}}_{k'}(\vec{\hat{x}}_{k'}^1,...,\vec{\hat{x}}_{k'}^d)\hat{\mathbf{f'}}_*(\{\vec{\hat{x}}_k^1,...,\vec{\hat{x}}_k^d\}_{k=1}^t)\right)]\}[i,j]\\
		&=\textup{F}^{-1}\{\textup{conj}(\textup{F}\{R'(x_{k'};f'_*)\})\}[i,j]\\
		&=\textup{conj}(R'(x_{k'};f'_*)[-i,-j])\\
		&=R'(x_{k'};f'_*)[-i,-j],\quad i\in [-\frac{m}{2},\frac{m}{2}], j\in [-\frac{n}{2},\frac{n}{2}],
	\end{split}
\end{equation}
where $\textup{F}^{-1}$ denotes the inverse Fourier transform, $\textup{vec}^{-1}$ is the inverse operation of the vectorization. Additionally, $y[i,j]=y[-i,-j]$ since y is a centrosymmetric Gaussian function. Therefore,
\begin{small}
	\begin{equation}\label{Equv_conv_corre_Eq12}
		\begin{split}
			&\|R(x_{k'};f_*)]-y\|^2\\
			&=\sum_{i=-\frac{m}{2}}^{\frac{m}{2}}\sum_{j=-\frac{n}{2}}^{\frac{n}{2}}\textup{conj}(R(x_{k'};f_*)[i,j]-y[i,j])(R(x_{k'};f_*)[i,j]-y[i,j])\\
			&=\sum_{i=-\frac{m}{2}}^{\frac{m}{2}}\sum_{j=-\frac{n}{2}}^{\frac{n}{2}}\textup{conj}(R'(x_{k'};f'_*)[-i,-j]-y[i,j])(R(x_{k'};f_*)[i,j]-y[i,j])\\
			&=\sum_{i=-\frac{m}{2}}^{\frac{m}{2}}\sum_{j=-\frac{n}{2}}^{\frac{n}{2}}(R'(x_{k'};f'_*)[-i,-j]-y[i,j])(R(x_{k'};f_*)[i,j]-y[i,j])\\
			&=\sum_{i=-\frac{m}{2}}^{\frac{m}{2}}\sum_{j=-\frac{n}{2}}^{\frac{n}{2}}(R'(x_{k'};f'_*)[-i,-j]-y[i,j])(\textup{conj}(R'(x_{k'};f'_*)[-i,-j])-y[i,j])\\
			&=\sum_{i=-\frac{m}{2}}^{\frac{m}{2}}\sum_{j=-\frac{n}{2}}^{\frac{n}{2}}(R'(x_{k'};f'_*)[-i,-j]-y[-i,-j])(\textup{conj}(R'(x_{k'};f'_*)[-i,-j])-y[-i,-j])\\
		\end{split}
	\end{equation}
\end{small}
\begin{small}
	\begin{equation}\label{Equv_conv_corre_Eq120}
		\begin{split}
			&=\sum_{i=-\frac{m}{2}}^{\frac{m}{2}}\sum_{j=-\frac{n}{2}}^{\frac{n}{2}}(R'(x_{k'};f'_*)[i,j]-y[i,j])(\textup{conj}(R'(x_{k'};f'_*)[i,j])-y[i,j])\\
			&=\sum_{i=-\frac{m}{2}}^{\frac{m}{2}}\sum_{j=-\frac{n}{2}}^{\frac{n}{2}}(R'(x_{k'};f'_*)[i,j]-y[i,j])\textup{conj}(R'(x_{k'};f'_*)[i,j]-y[i,j])\\
			&=\|R'(x_{k'};f'_*)-y\|^2.
		\end{split}
	\end{equation}
\end{small}

If $[(\sum_{k=1}^{t}\alpha_k \hat{\mathbf{X}}_k^\textup{H}\hat{\mathbf{X}}_k)+\lambda \textup{I}_{dN} ]$ is invertible, then $[(\sum_{k=1}^{t}\alpha_k \hat{\mathbf{X'}}_k^\textup{H}\hat{\mathbf{X'}}_k)+\lambda \textup{I}_{dN} ]$ is also invertible and vice versa, because they are conjugated. Based on the above deductions, the following conclusions can be drawn. 
\begin{proposition} \label{proposition1}
	If the optimal solution $\hat{\mathbf{f_*}}(\{\vec{\hat{x}}_k^1,...,\vec{\hat{x}}_k^d\}_{k=1}^t)$ in (\ref{eq:DCF_rep}) exists, then the optimal solution $\hat{\mathbf{f'_*}}(\{\vec{\hat{x}}_k^1,...,\vec{\hat{x}}_k^d\}_{k=1}^t)$ in (\ref{Equv_conv_corre_Eq1}) exists as well, and vice versa. Moreover, if $y$ is a 2-D centrosymmetric Gaussian function, then  
	\begin{equation}\label{Equv_conv_corre_Eq13}
		\begin{split}
			&\hat{\mathbf{f_*}}(\{\vec{\hat{x}}_k^1,...,\vec{\hat{x}}_k^d\}_{k=1}^t)=\textup{conj}(\hat{\mathbf{f'_*}}(\{\vec{\hat{x}}_k^1,...,\vec{\hat{x}}_k^d\}_{k=1}^t)).
		\end{split}
	\end{equation}
	Meanwhile, given a detection sample $x_{k'}$, the correlation filter response of $f_*=\textup{vec}^{-1}[\textup{F}^{-1}\{\hat{\mathbf{f_*}}\}]$ and the convolution filter response of $f'_*=\textup{vec}^{-1}[\textup{F}^{-1}\{\hat{\mathbf{f'_*}}\}]$, respectively, to $x_{k'}$ satisfy the following equations:
	\begin{align}
		&\|R(x_{k'};f_*)]-y\|^2=\|R'(x_{k'};f'_*)-y\|^2,\\
		R(x_{k'};f_*)[i,j]=&R'(x_{k'};f'_*)[-i,-j],\quad i\in [-\frac{m}{2},\frac{m}{2}], j\in [-\frac{n}{2},\frac{n}{2}].\\ \nonumber
	\end{align}
\end{proposition} 
Proposition (\ref{proposition1}) in fact shows that in visual tracking the correlation filter and the convolution filter are equivalent in the sense of equal minimum mean-square error of estimation, under the condition that the ideal filter response is a 2-D centrosymmetric Gaussian function and the optimal solutions exist. More specifically,

\begin{itemize}
	\item [*]As long as the optimal solution of one of the filters is known, the optimal solution of the other can be obtained immediately;
	\item [*]The mean-square errors of the filter responses of the two optimal solutions to a given detection sample are equal;
	\item [*]The filter responses of the two optimal solutions to a given detection sample are symmetric about the origin. Intuitively, assuming there is only one maximal value in the filter response of either the optimal solution to a given detection sample, when the coordinate of the maximal value of the filter response of one of the filters is estimated correctly, the other is also estimated correctly; when the coordinate of the maximal value of one of the filters deviates from the correct position, the coordinate of the maximal value of the other deviates from the correct position in the opposite direction, but their distances from the correct position are equal.
\end{itemize}

\section{Conclusion}

In this paper, for the first time we proves that correlation filter and convolution filter are equivalent in the sense that their minimum mean-square errors (MMSEs) in visual tracking are equal, under the condition that the optimal solutions exist and the ideal filter response is a 2-D centrosymmetric Gaussian function. This result gives researchers the freedom to choose correlation or convolution in formulating their trackers. It also suggests that the explanation of the ideal response in terms of similarities is not essential.

%
%
%
 \bibliographystyle{splncs04}
 \bibliography{sample}
\end{document}